\documentclass{article}

\usepackage[letterpaper,top=2cm,bottom=2cm,left=3cm,right=3cm,marginparwidth=1.75cm]{geometry}
\newcommand{\probP}{\text{I\kern-0.15em P}}
\usepackage[english]{babel}
\usepackage[autostyle]{csquotes}
\usepackage{amsmath}
\usepackage{graphicx}
\usepackage{mathtools}
\usepackage{amssymb}
\usepackage{bbm}
\usepackage{subfiles}%subfile
\usepackage{authblk}
\usepackage{fancyhdr}
\usepackage[dvipsnames]{xcolor}
\usepackage[document]{ragged2e}
\usepackage{lipsum}
\usepackage[linesnumbered,ruled,vlined]{algorithm2e}
\usepackage[colorlinks]{hyperref}
\hypersetup{
  colorlinks,
  citecolor=blue,
  linkcolor=red}
\usepackage[nameinlink,noabbrev]{cleveref} % always load cleveref *after* hyperref
\usepackage[square,numbers,sort&compress]{natbib}
\usepackage{subcaption}
\usepackage{multirow}
\usepackage{float}
\usepackage{amsthm}
\usepackage{multirow}
\usepackage{pdflscape}
\usepackage{threeparttable}
\crefalias{AlgoLine}{line}
\makeatletter
\renewcommand{\maketitle}{\bgroup\setlength{\parindent}{0pt}
\begin{flushleft}
  \textbf{\flushleft\LARGE\@title}
  \vspace{0.5cm}
  
  \@author
\end{flushleft}\egroup
}
\makeatother
\providecommand{\keywords}[1]
{
  \small	
  \textbf{\textit{Keywords:}} #1
}

\title{A Multivariate Bernoulli-Based Sampling Method for Multi-Label Data with Application to Meta-Research}
\author[1,2]{Simon Chung}
\author[3]{Colby J. Vorland}
\author[4]{Donna L. Maney}
\author[1,2,*]{Andrew W. Brown}
\affil[1]{Department of Biostatistics, University of Arkansas for Medical Sciences, Little Rock, AR}
\affil[2]{Arkansas Children's Research Institute, Little Rock, AR}
\affil[3]{Department of Epidemiology and Biostatistics, Indiana University School of Public Health-Bloomington, Indiana, IN}
\affil[4]{Department of Psychology, Emory University, Atlanta, GA}
\date{}
\begin{document}
\footnotetext[1]{Corresponding author}
\maketitle
Datasets may contain observations with multiple labels. If the labels are not mutually exclusive, and if the labels vary greatly in frequency, obtaining a sample that includes sufficient observations with scarcer labels to make inferences about those labels, and which deviates from the population frequencies in a known manner, creates challenges. In this paper, we consider a multivariate Bernoulli distribution as our underlying distribution of a multi-label problem. We present a novel sampling algorithm that takes label dependencies into account. It uses observed label frequencies to estimate multivariate Bernoulli distribution parameters and calculates weights for each label combination. This approach ensures the weighted sampling acquires target distribution characteristics while accounting for label dependencies. We applied this approach to a variety of datasets, including a sample of research articles from Web of Science labeled with 64 biomedical topic categories. We aimed to preserve category frequency order, reduce frequency differences between most and least common categories, and account for category dependencies. This approach produced a more balanced sub-sample, enhancing the representation of minority categories.
\vspace{0.25 cm}


\section{Introduction}

In certain real-world applications such as sampling scientific articles in meta-research, datasets may contain observations with multiple labels that are not necessarily mutually exclusive \citep{ZHANG2014}. For example, a scientific article in a bibliometric database might be assigned several categories that reflect different disciplines (e.g., immunology, neurosciences, psychiatry). Because these labels are linked to the same observation, the dataset cannot be treated as a collection of independent categories, and any data manipulation targeting one specific label inherently alters the marginal distributions of the others. Sampling or modeling from datasets with a multi-label structure is part of multi-label learning, where the goal is to obtain information from such datasets. Additionally, some labels may have very low frequencies, making it difficult to support reliable training, evaluation, or inference \citep{CHARTE2015A, LIU2022}. Attempting to characterize rare labels may lead to misleading or non-generalizable results. Under an imbalanced distribution, we are likely to obtain a sample that inherits these imbalance characteristics using simple random sampling and smaller samples might lack of observations with rare labels \citep{SECHIDIS2011}. We therefore may not obtain a sufficient number of observations on the rare labels to permit inferences. A different approach is needed to ensure that we can make reliable inferences both at the label level and for the dataset as a whole. A similar concern in machine learning is commonly known as multi-label classification (MLC) \citep{TSOUMAKAS2010}, where the main challenge lies in the imbalanced distribution across datasets \citep{KAUR2019, HERRERA2016, CHARTE2015B, DANIELS2017, LIU2018}. Given the popularity of MLC in recent years, most of the algorithms \citep{LIU2022, CHARTE2015B, SADHUKHAN2019, PEREIRA2020, CHARTE2019} developed for multi-label learning (MLL) \citep{HERRERA2016} have focused on the supervised learning setting, where they assume access to a set of input features and treat the labels as target variables to be predicted. Some algorithms pre-select or generate samples for learning \citep{CHARTE2014, CHARTE2019, GENG2016} while the others make the classifier more robust to an imbalanced label distribution \citep{SUN2007, ZHANG2020, DANIELS2017, LIU2018, HAN2005, TAHIR2012, RASTIN2021}. Yet, in meta-research, where scientific articles are routinely sampled, the labels (e.g., study characteristics, research topics) are often used to describe the studies themselves, and are highly imbalanced, making the data structure similar to the MLC setting but with a fundamentally different goal --- characterizing rather than predicting.

In addition to label imbalance, a central challenge in sampling from multi-label datasets is the inherent dependency structure among labels \citep{WANG2021, LI2019, LIU2018}. Unlike in a single-label setting, where labels are mutually exclusive, labels in multi-label data often co-occur in complex and non-independent ways, posing a major obstacle for traditional sampling methods. In a simpler scenario, in which label dependency is absent, the problem reduces to multi-class sampling in which each observation in our data can only have one label. Common approaches to address imbalanced sampling include oversampling and undersampling by down-weighting the probability of sampling common labels (i.e., making selection of observations with common labels less likely, or \enquote{undersampling} them relative to the population) or up-weighting uncommon labels (i.e., making selection of observations with uncommon labels more likely, or \enquote{oversampling} them)\citep{BARANDELA2003, CHAWLA2002, HAN2005, KOTSIANTIS2006, MOHAMMED2020, GARCIA2012}. When each label is mutually exclusive, this can be done simultaneously through approaches such as inverse frequency weighting. When label dependency exists, such an approach can lead to increasing the imbalanced level for other labels \citep{LIU2019}, when observations can have multiple labels. 

Most existing algorithms for sampling multi-label data, heavily rely on metrics like Imbalance Ratio (IR) \citep{CHARTE2013} to assess the imbalance level. While these metrics provide useful heuristics for addressing label imbalance, the algorithms typically operate without an explicit distribution model. As a result, the sampling decision is not guided by the underlying data-generating process but by label frequency. Furthermore, since these algorithms typically yield deterministic outcomes when presented with a fixed dataset and set of hyperparameters, they might not possess the stochasticity required in a sampling procedure that consistently returns the same samples across multiple runs. Beyond selection-based strategies, some algorithms focus on synthesizing new data from the minority samples. Yet, this reliance on numerical interpolation poses significant challenges for discrete data such as in text samples, where the complex nature of the feature space poses significant challenges for constructing valid intermediate samples. The main contributions of this paper are therefore twofold:
\begin{itemize}
\item We propose a novel, probabilistically grounded sampling algorithm designed for general-purpose multi-label learning. This model-agnostic approach operates exclusively within the label space, functioning independently of specific downstream MLC architectures or predictor variables.
\item We introduce an advanced sampling framework that strategically preserves some level of imbalance when perfect label equilibrium is either mathematically intractable or practically undesirable, allowing for a controlled level of natural imbalance.
\end{itemize}

The rest of this paper is organized as follows. In \cref{sec:related}, we provide a review of the literature relevant to our algorithms. \Cref{sec:algorithm} outlines the assumptions regarding the distribution of labels and defining notations. Subsequently, it introduces algorithms tailored to different sampling requirements. The first application, which serves as the motivation for the algorithm, is detailed in \cref{sec:application}. Additionally, \cref{sec:discussion} discusses the challenges encountered during the algorithm's execution and the similarity between our algorithm and inverse probability weighting. \Cref{sec:conclusion} summarizes our work.

\section{Related Work}\label{sec:related}

Early sampling strategies for multi-label datasets adapted the same ideas used for dealing with multi-class datasets. Approaches such as LP-based random undersampling/oversampling (LP-RUS \& LP-ROS) treat each unique labelset as a distinct class, and individual label random undersampling/oversampling (ML-RUS \& ML-ROS) shifted the focus to evaluating the imbalance level for each individual label \citep{CHARTE2015A}. 

Given that simple undersampling results in a loss of information and simple oversampling results in overfitting, heuristic approaches were integrated into subsequent algorithms. Many of these metric-based sampling algorithms employ heuristic strategies that iterate through all samples in the dataset, evaluating the effect of including or excluding each observation on the overall imbalance.  For undersampling, MultiLabel edited Nearest Neighbor (MLeNN) \citep{CHARTE2014} was introduced as a heuristic strategy that removes samples that do not belong to the minority labels. Similarly, Multi-Label Tomek Link (MLTL) \citep{PEREIRA2020} utilized undersampling by removing majority label instances around the minority class using Tomek links, and the Multi-Label UndersampLing method (MLUL) \citep{LIU2022} heuristically removes \enquote{harmful} instances that interfere with the learning of their reverse k-nearest neighbors. Partial random undersampling and oversampling (PRUS \& PROS) \citep{GARCIA-PEDRAJAS2024} explore the possibility of choosing different data for different classifiers to predict each label, preventing the loss of information by using a mask to select which instances are used for training.

Over the years, results have shown that global imbalance metrics are insufficient to improve classifier performance because the true difficulty of a classifier lies in the overlapping of classes within local neighborhoods. This realization led to algorithms focused on synthesizing new data from minority samples. MultiLabel synthetic instance generation (MLSMOTE) \citep{CHARTE2015B} integrated SMOTE \citep{CHAWLA2002} with imbalance ratios. The Multi-Label Synthetic Oversampling approach based on the Local distribution of labels (MLSOL) \citep{LIU2019, LIU2022} was proposed to synthesize datapoints specifically near hard-to-learn examples based on local distributions. Multi-Label Borderline Oversampling Technique (MLBOTE) \citep{TENG2024} further categorizes borderline samples into self-borderline and cross-borderline that impact the global and local imbalance levels, while Diversity and Reliability-enhanced SMOTE (DR-SMOTE) \citep{GONG2025} addresses MLSMOTE's tendency to generate noisy and unreliable synthetic data. \citeauthor{ren2025} \citep{ren2025} proposed MLCIO, a multi-label over-sampling method that uses rarity-weighted voting rather than majority voting to assign labels to synthesized samples, preventing minority labels from being outvoted by co-occurring majority labels.

Recently, some researchers have started to move beyond heuristic selection and interpolation. \citeauthor{CHO2025} \citep{CHO2025} proposed categorizing labels using empirical frequencies and percentiles. However, the threshold of rare and common labels, which are less than $25^\text{th}$ and  greater than $75^\text{th}$ percentile of the frequency distribution respectively, and the resampling strategy corresponding to each label class, are somewhat arbitrary and without sufficient theoretical background. \citeauthor{ZHOU2025} \citep{ZHOU2025} used sample uncertainty derived from the model prediction using entropy and label correlation to guide the sample selection process; nevertheless, the post-sample distribution tied directly to how well the model learns and predicts, which does not necessarily lead to label balance.

Driven by the increasing scale of modern datasets and the high cost of exhaustive annotation, much of the recent multi-label literature remains tied to classifier-dependent frameworks, ranging from pure supervised models to semi-supervised and partial learning approaches. Multi-label Learning based on Hierarchical Clustering (MLHC) \citep{DUAN2024} used hierarchical clustering to group labels based on their correlation and transform the multi-label problem into multi-class problem. Clustering keeps each transformed subspace's class count manageable while preserving genuine label correlation structure, addressing a key limitation of prior methods that ignored or randomly modeled label correlations. Co-Label Selection (CLS) \citep{lyu2025} shift the strategy from sample selection to label selection, using two peer networks that exclude individual negative labels from back-propagation rather than dropping entire samples. A two-step filtering process removes likely false-negative labels and then easy, uninformative negative labels to address label imbalance.

In summary, while the field has evolved from global sampling and heuristic selection to more theory grounded frameworks, a critical methodological gap describing the distribution behind the labels remains. Existing advanced attempts either rely on arbitrary thresholds that lack sufficient theoretical foundation, or use selection strategies strictly tied to a specific classifier, solving only a single problem rather than addressing the underlying imbalance problem. These limitations underscore the pressing need for a principled, mathematically rigorous framework that directly models the multivariate label dependencies and co-occurrence structures from a sound theoretical basis, serving not just supervised learning tasks, but offering a more general-purpose solution.

\section{Proposed approach}\label{sec:algorithm}

\subsection{Indexing}
We introduce mappings between any $K$-dimension binary vectors $\begin{bmatrix} y_{1} & \dots & y_{K}\end{bmatrix}$ to integers. By seeing $y$ as a binary digit, we can map it to an integer with the following function:
$$\mathcal{J}:\{0,1\}^K\mapsto\{0,1,\dots, 2^K-1\},\mathcal{J}(y) = \mathcal{J}(y_1,\dots,y_K)=\sum\limits_{i=1}^{K}2^{(i-1)}\cdot(y_i-1).$$
Conversely, we can also map an integer back to $K$-dimension space as following:
$$\mathcal{J}^{-1}:\left\{0,1,\dots, 2^K-1\right\}\mapsto\{0,1\}^K, \mathcal{J}^{-1}(z) = \left[\left\lfloor \frac{z}{2^{i-1}} \right\rfloor\text{ mod } 2\right]_{i\leqslant K}.$$
This mapping arranges all $2^K$ variables along a one-dimensional sequence. Herein, any variable indexed by the binary vector $(y_1,\dots,y_K)$ will be ordered according to $\mathcal{J}(y_1,\dots,y_K)$.  
\subsection{Definition}
Suppose there are $K$ labels $\{\mathcal{A}_1,\dots,\mathcal{A}_K\}$ in the dataset, $N$ observations in our sample, and $S_n$ is the labelset of $n^{\text{th}}$ observation. Let the $K$-dimensional random vector of labels $$Y^{(n)} = \begin{bmatrix} Y^{(n)}_{1} & \dots & Y^{(n)}_{K}\end{bmatrix}\in\{0,1\}^K,\forall n\leqslant N$$ be the indicator vector of $S_n$, where $Y^{(n)}_i = 1$ if the $i^{\text{th}}$ label is in the labelset $S_n$ and $Y^{(n)}_i = 0$ if the $i^{\text{th}}$ label is not in the labelset $S_n$, i.e.
\begin{equation*}
Y^{(n)}_i= I\left(\mathcal{A}_i \in S_n\right) = 
\begin{cases}
  1 & \mathcal{A}_i \in S_n,\\
  0 & \mathcal{A}_i \notin S_n.
\end{cases},\forall i\leqslant K,n\leqslant N,
\end{equation*}
where $I$ is the identity function. We can also refer the indicator vector of  $i^{\text{th}}$ labelset as $\mathcal{J}^{-1}(i)$ and $k^{\text{th}}$ label as $Y_k$. In this paper, we assume $Y$ follows a multivariate Bernoulli (MVB) distribution \citep{DAI2013}, i.e.
$$Y = \left(Y^{(1)},\dots,Y^{(N)}\right) \overset{i.i.d}{\sim} \text{MVB}\left(p\right),$$
where $p := \left[p_{y}\right]_{y\in\{0,1\}^K}$ is the parameter set of the MVB and labelset indicator $y:=\begin{bmatrix} y_{1} & \dots & y_{K}\end{bmatrix}$ is the realization of $Y$.  Here, we only consider the case where each observation has at least one label, which means that $p_{0,0,\dots,0}$ is 0. Then $Y^{(n)}$ has the following probability mass function by \citep{DAI2013}:
\begin{equation*}
\probP(Y=y) = \probP(Y_1 = y_1, \dots, Y_K = y_K) = p_{1,0,\dots,0}^{\left[y_1\prod\limits_{i=2}^{K} \left(1-y_i\right)\right]}\cdot p_{0,1,\dots,0}^{\left[\left(1-y_1\right)y_2\prod\limits_{i=3}^{K} \left(1-y_i\right)\right]}\cdots p_{1,\dots,1}^{\left[\prod\limits_{i=1}^{K} y_i\right]}.
\end{equation*}
Hence, we can obtain the marginal distributions as 
\begin{equation*}
\probP(Y_1 = y_1, \dots, Y_r = y_r) 
=\sum\limits_{y_{r+1}}\cdots\sum\limits_{y_{K}}p_{y_1,\dots,y_K}, \forall r\leqslant K.
\end{equation*}

We will use the empirical distribution of our sample to estimate the population parameters. Suppose there are $N$ observations in our sample. Let $y^{(n)} = \begin{bmatrix} y^{(n)}_{1} & \dots & y^{(n)}_{K}\end{bmatrix}$ be the indicator vector of labels of the $n^{\text{th}}$ observation. Then we can have
$$\hat p_{y} = \frac{\sum\limits_{n=1}^{N}I(y^{(n)}=y)}{N}.$$ 
This empirical estimate $\hat{p}_{y}$ is also the maximum likelihood estimator (MLE) of the parameters of MVB. By the properties of MLE, $\hat p_y$ is a consistent estimator of the true population parameter (\nameref{sec:appendix}).
\subsection{Balanced Sampling}\label{subsec:balanced_sampling}
In our first use case, we want the distribution in the subsample $Y^\prime$ we draw to be balanced across each label. We aimed to assign a weight $q_{y_1,\dots,y_K}$ to each labelset such that the marginal distribution of $Y^{'}_i$ in our sample distribution is the same for all $1\leqslant i \leqslant K$, i.e.
$$\probP(Y^{'}_1 = 1) = \probP(Y^{'}_2 = 1) = \cdots = \probP(Y^{'}_K = 1),$$
where
$$Y^{'}\sim \text{MVB}\left(\left[q_y\cdot p_y\right]_{y\in\{0,1\}^K}\right) \text{ and }$$
$$\probP(Y^{'}_i = 1) = \sum\limits_{y_1}\cdots\sum\limits_{y_{i-1}}\cdot\sum\limits_{y_{i+1}}\cdots\sum\limits_{y_{K}}q_{y_1,\dots,y_{i-1},1,y_{i+1},\cdots,y_K}\cdot p_{y_1,\dots,y_{i-1},1,y_{i+1},\cdots,y_K}.$$

In order for $\left[q_y\cdot p_y\right]_{y\in\{0,1\}^K}$ to be a valid parameter set, $q_{y}$ has to meet the following requirements:
\begin{enumerate}
	\item $$q_{y_1,\dots,y_K}\cdot p_{y_1,\dots,y_K}\in[0,1],$$
	\item $$ \probP(\Omega) = \sum\limits_{y_1}\cdots\sum\limits_{y_{K}}q_{y_1,\cdots,y_K}\cdot p_{y_1,\cdots,y_K}=1,$$
    where $\Omega$ is the sample space, in our case, the union of all the possible combination of $K$ labels,
    \item \begin{equation*}\label{eq:1}
    \begin{cases}
    \probP(Y^{'}_1 = 1) = \sum\limits_{y_2}\cdots\sum\limits_{y_{K}}q_{1,y_2,\cdots,y_K}\cdot p_{1,y_2,\cdots,y_K} = b\\
    \probP(Y^{'}_2 = 1) = \sum\limits_{y_1}\cdot\sum\limits_{y_3}\cdots\sum\limits_{y_{K}}q_{y_1,1,\cdots,y_K}\cdot p_{y_1,1,\cdots,y_K} = b\\
    \vdots\\
    \probP(Y^{'}_K = 1) = \sum\limits_{y_1}\cdots\sum\limits_{y_{K-1}}q_{y_1,\cdots,y_{K-1},1}\cdot p_{y_1,\cdots,y_{K-1},1} = b\\
    \end{cases}, \text{ where } b\in[0,1].
    \end{equation*}
\end{enumerate}
For \ref{eq:1}, we can rewrite it in a matrix format for solving as a system of linear equations $$A\cdot Q = b1_{K\times1},$$
where 
\begin{equation*}
\begin{aligned}
	&A =
\begin{bmatrix}
 p_{1,0,0,\dots,0} \cdot 1 & p_{0,1,0,\dots,0}\cdot 0 & p_{1,1,0,\dots,0} \cdot 1 & \cdots & p_{1,1,1,\dots,1} \cdot 1\\
 p_{1,0,0,\dots,0} \cdot 0 & p_{0,1,0\dots,0}\cdot 1 & p_{1,1,0,\dots,0} \cdot 1 & \cdots & p_{1,1,1,\dots,1} \cdot 1 \\
 \vdots & \vdots & \vdots & \ddots & \vdots \\
 p_{1,0,0,\dots,0} \cdot  0 & p_{0,1,0\dots,0} \cdot 0 & p_{0,1,0\dots,0} \cdot 0 &\cdots & p_{1,1,1,\dots,1} \cdot 1
\end{bmatrix},
&Q = \begin{bmatrix}
q_{1,0,0,\dots,0}\\
q_{0,1,0,\dots,0}\\
\vdots\\
q_{1,1,1,\dots,1}
\end{bmatrix},	
\end{aligned}
\end{equation*}
which is equivalent as
\begin{equation*}
	\begin{bmatrix}
	A & | & -1
	\end{bmatrix}\cdot	
	\begin{bmatrix}
	Q\\
	-\\
	b
	\end{bmatrix}
	=0_{K\times 1}.
\end{equation*}
Therefore, we can summarize these characteristics into an optimization problem to solve for $q$, i.e.
\begin{equation*}\label{eq:2}
\begin{aligned}
	&\text{min} &&|AQ-b1_{K\times1}|\\
	&\text{subject to} &&PQ=1,\\
	& &&\text{where } P =\begin{bmatrix}p_{0,0,0,\dots,0} & p_{1,0,0,\dots,0} & p_{0,1,0,\dots,0} & \cdots & p_{1,1,1,\dots,1}\end{bmatrix}&&&\text{Equality Constraint}\\
    & &&0\leqslant q_{y_1,y_2,\dots,y_K}\leqslant \frac{1}{p_{y_1,y_2,\dots,y_k}} &&&\text{Inequality Constraint}\\
	& &&0\leqslant b\leqslant1 &&&\text{Inequality Constraint}\\
\end{aligned}
\end{equation*}
Let $q^\star$ be the solution of the above constrained optimization. We denote this optimization step compactly as 
$$q^\star = \mathcal{O}(p)$$ as
$\mathcal{O}(\cdot)$ is defined as the constrained optimizer with respect to the feasible set. Then we can assign the $q^\star$ to the observation in our sample. The subsample from the weighted sampling with $q^\star$ will follow MVB with equal marginal distribution.

\begin{algorithm}
  \caption{MVB Based Balanced Sampling}
  \label{alg:1}
  \textbf{Data}: $S = \{S_n\subset\{\mathcal{A}_1,\dots,\mathcal{A}_K\}\mid n\leqslant N\}$, \textbf{Subsample Size}: $m$;
  
  Compute the corresponding indicator $y:=\{y^{(n)}=\left[I(\mathcal{A}_i\in S_n)\right]_{i\leqslant K} \in \{0,1\}^{K}\mid n\leqslant N\}$ ;

  Compute the empirical distribution $\hat p = \left[\frac{1}{N}\sum\limits_{n=1}^N I(y^{n}=y)\right]_{y\in \{0,1\}^K}$;

  $\hat{q^\star} = \mathcal{O}(\hat p)$;

  Assign weights to observations $\hat Q = \{\hat{q^\star}_{y^{(n)}}\mid n\leqslant N\}$;

  $\hat y \sim \text{Weighted Sampling} (y, \hat Q, m)$

\end{algorithm}
\subsection{Compressed Imbalanced Sampling}\label{subsec:compressed_sampling}
Suppose we want the marginal distributions of the sample to preserve the most of order of frequency from the population marginal distribution, but to reduce the imbalance across the labels. One way is to compress the relative ratio between each marginal distribution. Let
\begin{equation*}
	P_M = \begin{bmatrix}\probP(Y_1 = 1) & \probP(Y_2 = 1) &\cdots& \probP(Y_K = 1)\end{bmatrix}^T
\end{equation*}
be the vector of marginal probability of $Y$.
Our goal is to compress the ratio between the most and least common labels, using a function with known properties that maintains the relation among labels and their dependencies. Hence, the ratio before sampling can be expressed as
\begin{equation*}
	R = \frac{P_M}{\max{P_M}}.
\end{equation*}
After compressing, the ratio between categories can be
\begin{equation*}
	R^\prime = \sqrt[s]{R},
\end{equation*}
where $s$ is the strength of the compressing and $R^\prime\to\begin{bmatrix}1&\cdots&1\end{bmatrix}^T$ as $s\to\infty$, which will be equivilant to Balanced Sampling from \cref{subsec:balanced_sampling}.

In addition, our marginal distribution after sampling will be $bR^\prime$, which leads to the optimization algorithm
\begin{equation*}\label{eq:3}
\begin{aligned}
	&\text{min}&&|AQ-bR^\prime|\\
	&\text{subject to} &&PQ=1\\
	& &&\text{where } P =\begin{bmatrix}p_{0,0,0,\dots,0} & p_{1,0,0,\dots,0} & p_{0,1,0,\dots,0} & \cdots & p_{1,1,1,\dots,1}\end{bmatrix}&&&\text{Equality Constraint}\\
    & &&0\leqslant q_{y_1,y_2,\dots,y_K}\leqslant \frac{1}{p_{y_1,y_2,\dots,y_k}} &&&\text{Inequality Constraint}\\
	& &&0\leqslant b\leqslant1 &&&\text{Inequality Constraint}\\
\end{aligned}
\end{equation*}

Similarly, we let $q^\star$ be the solution of the above constrained optimization and the optimization step as 

$$q^\star = \mathcal{O}_s(p, R^\prime)$$

as $\mathcal{O}$ defined as the constrained optimizer in compressed imbalanced sampling. Similarly, with the optimal solution $q^\star$, the subsample will follow the target distribution with compressed relative ratio of the marginal distribution.

\begin{algorithm}
  \caption{MVB Based Compressed Imbalanced Sampling}
  \label{alg:2}
  \textbf{Data}: $S = \{S_n\subset\left\{\mathcal{A}_1,\dots,\mathcal{A}_K\right\}\mid n\leqslant N\}$, \textbf{Subsample Size}: $m$, \textbf{Compressing Strength}: $s$;
  
  Compute the corresponding indicator $y:=\{y^{(n)}=\left[I(\mathcal{A}_i\in S_n)\right]_{i\leqslant K} \in \{0,1\}^{K}\mid n\leqslant N\}$ ;

  Compute the empirical distribution $\hat p = \left[\frac{1}{N}\sum\limits_{n=1}^N I(y^{n}=y)\right]_{y\in \{0,1\}^K}$;
  
  Compute $R^\prime$ based on $s$ and $\hat p$;

  $\hat{q^\star} = \mathcal{O}_s(\hat p, R^\prime)$;

  Assign weights to observations $\hat Q = \{\hat{q^\star}_{y^{(n)}}\mid n\leqslant N\}$;

  $\hat y \sim \text{Weighted Sampling} (y, \hat Q, m)$

\end{algorithm}

\section{Applications}\label{sec:application}

This algorithm was motivated by a common challenge in meta-research and bibliometric studies, in which researchers often aim to sample publications across a broad spectrum of research areas. In many of these applications, articles are tagged with categories, such as those assigned by Web of Science (WoS) \citep{WOS}, which serve as a designation for disciplinary or topical coverage. Because each article can belong to multiple categories, the collection naturally forms a multi-label dataset. Here, we describe sampling from two corpora, one with more than 200,000 articles and more than 60 categories, and a smaller one with only 6 categories. Our algorithm performs well in both cases.

\subsection{Example 1: Sampling from a large corpus with 64 categories} \label{sec:example1}

We obtained 221,400 articles from WoS labeled with 64 categories of interest published in 2023. Each article has at least one and can have more than one category. There is theoretically no upper bound to the number of categories an article can be labeled with, but in our set, the proportions of articles with 1, 2, 3, 4, or 5 labels were 55.8\%, 31.5\%, 10.1\%, 2.3\%, and 0.3\%, respectively. The histogram in \cref{fig:main_before} shows that the marginal distribution of categories in our corpus is not uniform. After random sampling, we can see from \cref{fig:main_RS} that the sample we obtained inherits the imbalanced characteristic from the corpus.
We aimed to systematically sample 3,000 articles from our corpus such that the sample will not be dominated by the majority categories while retaining a distribution of label frequencies ordinally similar to the original corpus, within sampling error. We applied MVB-based compressed imbalanced sampling (\cref{alg:2}) to our corpus with compressing strength $s=2$. The histogram in \cref{fig:main_AS} shows that the marginal distribution is flatter and the relative order of the categories in terms of frequency is mostly preserved.

\subsection{Example 2: Sampling from a smaller corpus with six categories} \label{sec:example2}

\begin{figure}[H]
\centering
% Row 1 (wider than textwidth)
\makebox[\textwidth][c]{%
  \begin{subfigure}[b]{0.333\textwidth}
    \centering
    \renewcommand{\thesubfigure}{a}
    \includegraphics[width=\linewidth]{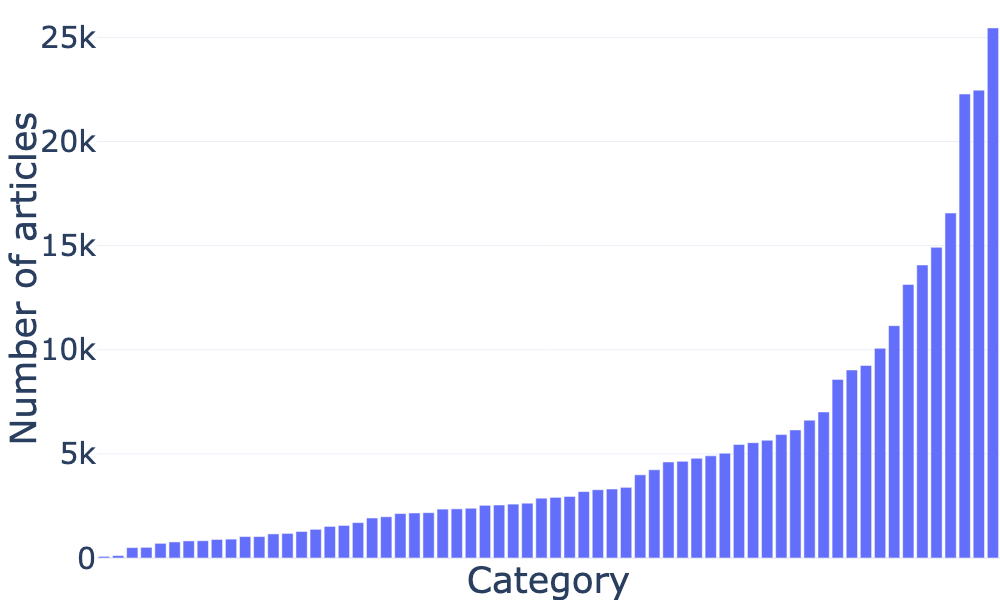}
    \caption{Example 1: Before Sampling}
    \label{fig:main_before}
  \end{subfigure}%
  \hspace{1em}%
  \begin{subfigure}[b]{0.333\textwidth}
    \centering
    \renewcommand{\thesubfigure}{b}
    \includegraphics[width=\linewidth]{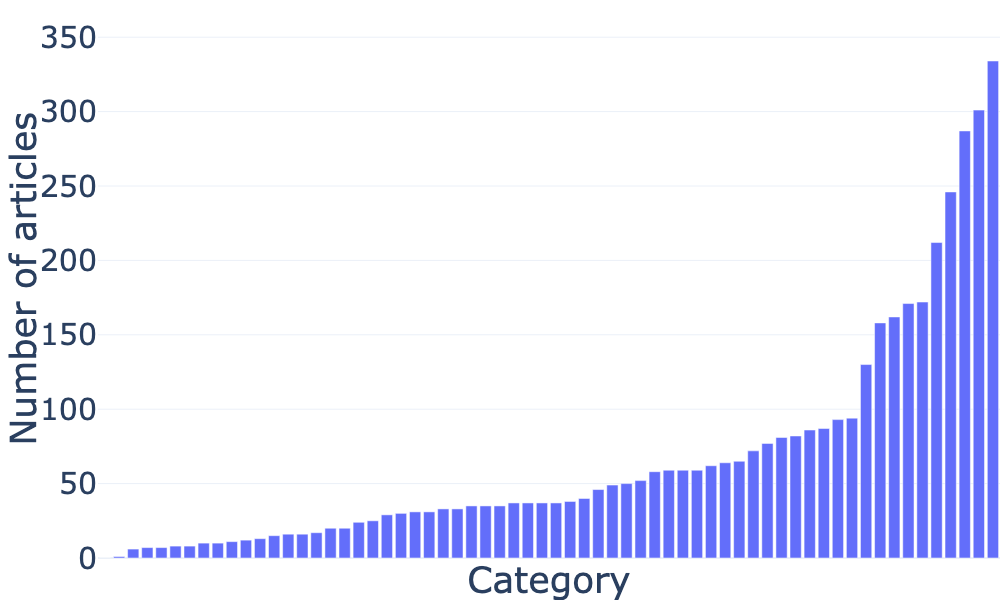}
    \caption{Example 1: Random Sampling}
    \label{fig:main_RS}
  \end{subfigure}%
  \hspace{1em}%
  \begin{subfigure}[b]{0.333\textwidth}
    \centering
    \renewcommand{\thesubfigure}{c}
    \includegraphics[width=\linewidth]{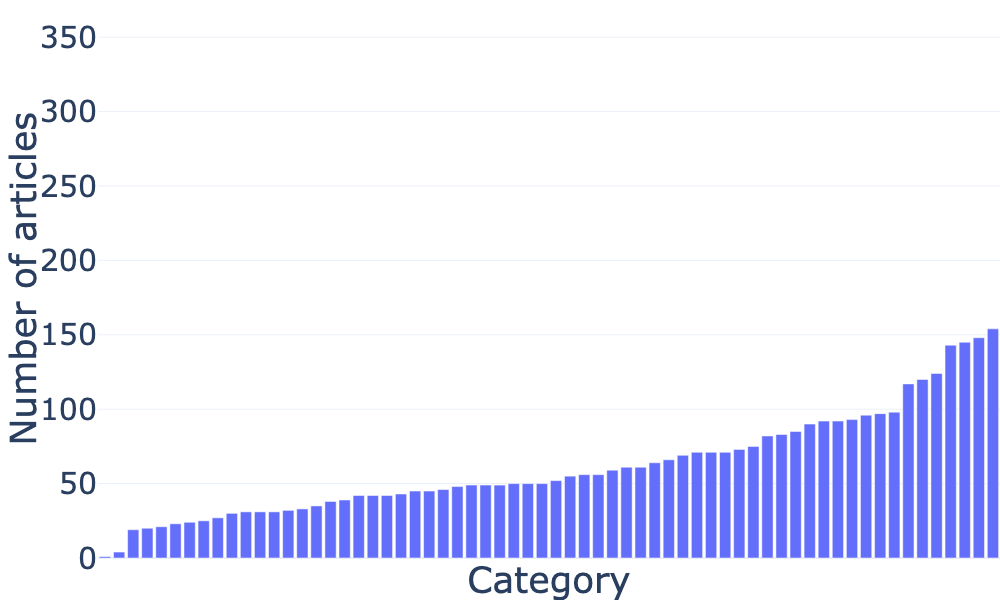}
    \caption{Example 1: MVB-based Sampling}
    \label{fig:main_AS}
  \end{subfigure}%
}

%\vspace{1em}

% Row 2
\makebox[\textwidth][c]{%
\begin{subfigure}[b]{0.333\textwidth}
    \centering
    \renewcommand{\thesubfigure}{d}
    \includegraphics[width=\linewidth]{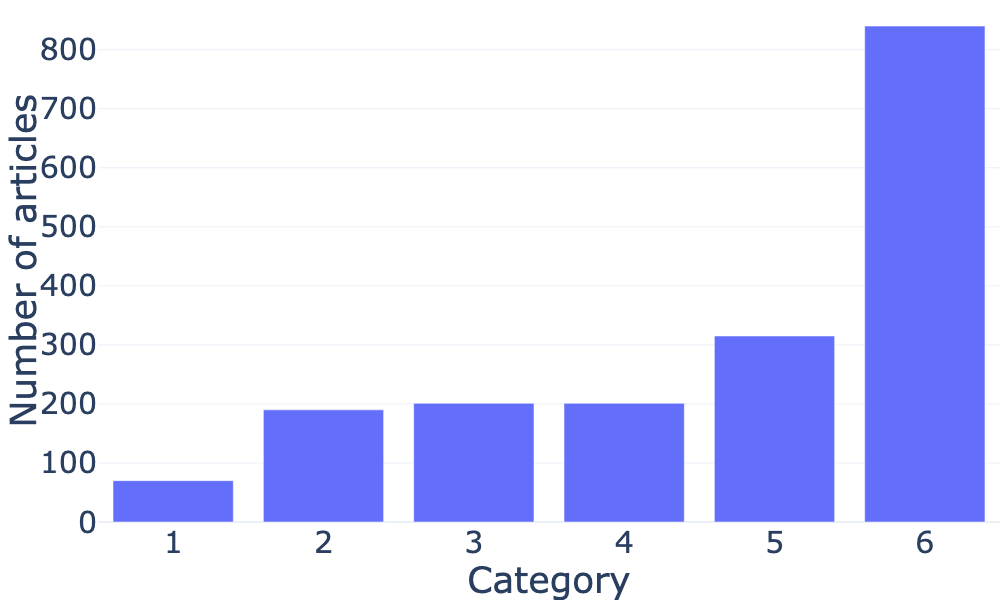}
    \caption{Example 2: Before Sampling}
    \label{fig:madi_before}
  \end{subfigure}%
  \hspace{1em}%
  \begin{subfigure}[b]{0.333\textwidth}
    \centering
    \renewcommand{\thesubfigure}{e}
    \includegraphics[width=\linewidth]{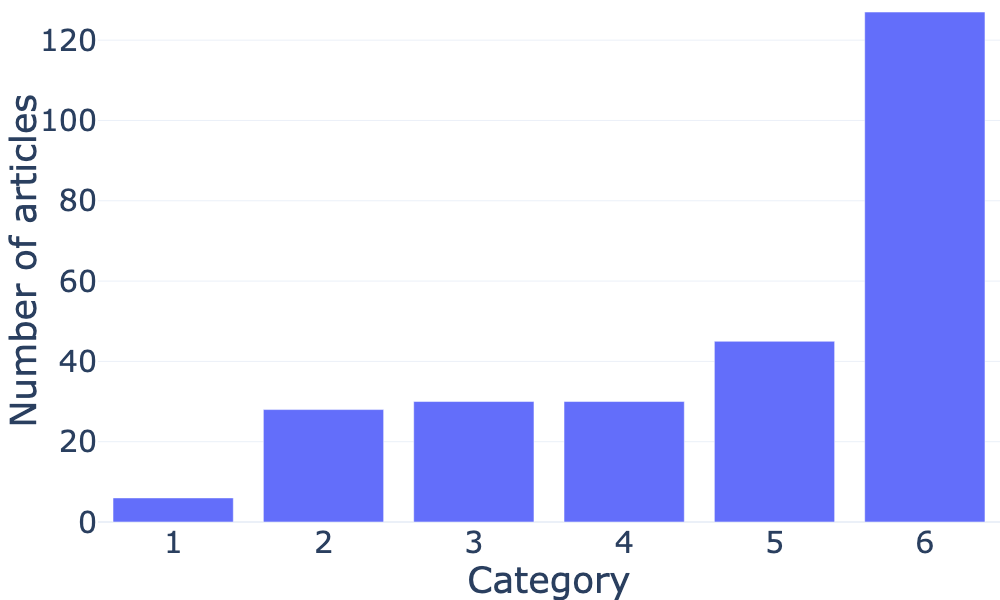}
    \caption{Example 2: Random Sampling}
    \label{fig:madi_RS}
  \end{subfigure}%
  \hspace{1em}%
  \begin{subfigure}[b]{0.333\textwidth}
    \centering
    \renewcommand{\thesubfigure}{f}
    \includegraphics[width=\linewidth]{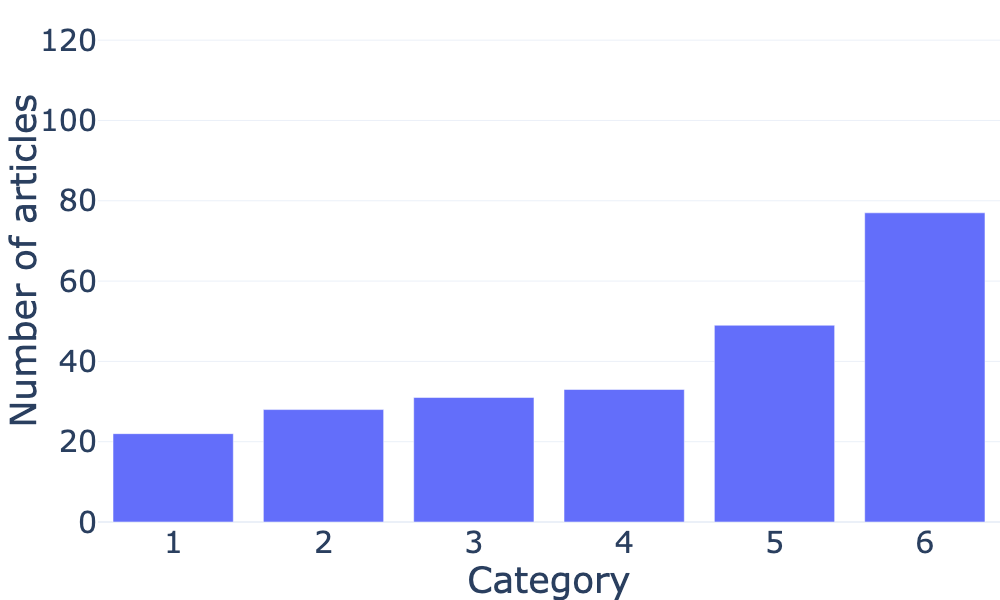}
    \caption{Example 2: MVB-based Sampling}
    \label{fig:madi_AS}
  \end{subfigure}%
}
\caption{Label distributions for two examples before sampling, after random sampling, and after MVB-based sampling. Categories on the x-axis are ordered by frequency in the original corpus. Note that the histogram will include more counts than total observations because of the existence of multiple labels.}
\label{fig:madi}
\end{figure}

In our second example corpus also from WoS, articles were published between 2019 and 2023 sampling 6 specific categories. We aimed to systematically sample 200 articles from 1,341 articles. As we can see from the marginal distribution in \cref{fig:madi_before}, this corpus also was characterized by a non-uniform distribution, which would be dominated by neural sciences (Category 6) with simple random sampling as shown in \cref{fig:madi_RS}. We applied MVB-based compressed imbalanced sampling (\cref{alg:2}) to this corpus using compressing strength $s = 2$ and obtained a more balanced sample, as the marginal distribution shows in \cref{fig:madi_AS}, which has been successfully demonstrated in recent work \citep{olivier2026}.

\subsection{Example 3: Datasets from an open-source repository} \label{sec:example3}
Even though our algorithms were developed to address sampling in meta-research, the usage can be extended to any dataset that is structured for multi-label learning. To validate our methods across different contexts, we applied our algorithm to 11 open-source datasets from a publicly available repository: \url{https://www.uco.es/kdis/mllresources/}. The performance of our algorithm on these datasets was measured using the average class imbalance ratio within each label (MeanIR) \citep{CHARTE2013}:
$$\text{MeanIR} = \frac{1}{K}\sum\limits_{k=1}^{K}\text{IR}_k,$$
where the $\text{IR}_k$ is the imbalance ratio per label (IRLbl) of the $k^{\text{th}}$ label $Y_k$:
$$\text{IR}_k=\frac{\max\limits_{1\leqslant k \leqslant K}\left\{\sum\limits_{n=1}^{N}Y^{(n)}_k\right\}}{\sum\limits_{n=1}^{N}Y^{(n)}_k},\forall k=1,2,\dots,K.$$
A higher IR indicates a more severely imbalanced dataset, with a value of 1 representing perfect balance.
\begin{landscape}
\begin{table}[htbp]
    \centering
    \begin{threeparttable}
    \captionsetup{justification=raggedright, singlelinecheck=false}
    \begin{tabular}{||l|c c c c c c c||}
    \hline
    % First row of headers
    \multirow{2}{*}{Dataset} & \multirow{2}{*}{Observations} & \multirow{2}{*}{Labels} & \multirow{2}{*}{Observed Labelsets} & \multicolumn{3}{c}{Imbalance Ratio (IR)} & \multirow{2}{*}{Time (s)} \\ 
    \cline{5-7} % Draws a horizontal line only across columns 5 and 7
    
    % Second row of headers
    & $N$ & $K$ & & Original & MVB Sampling (95\%  CI)\tnote{$\dagger$} & p-value\tnote{$\dagger$} & \\ [0.5ex] 
    \hline
    Image & 2000 & 5 & 20 & 1.193 & 1.031 [1.010, 1.061] & $<$ 0.001 & 0.009\\
    CHD 49 & 555 & 6 & 34 & 5.766 & 1.069 [1.025, 1.134] & $<$ 0.001 & 0.009\\
    Emotions & 593 & 6 &  27 & 1.478 & 1.078 [1.029, 1.149] & $<$ 0.001 & 0.008\\
    Flags & 194 & 7 & 54 & 2.255 & 1.097 [1.035, 1.189] & $<$ 0.001 & 0.010\\
    Yeast & 2417 & 14 & 198 & 7.197 & 1.044 [1.019, 1.081] & $<$ 0.001 & 0.018\\
    Water-quality & 1060 & 14 & 825 & 1.767 & 1.064 [1.034, 1.108] & $<$ 0.001 & 0.058\\
    Genbase\tnote{$\ast$} & 662 & 27 & 32 & 37.315 & 3.242 [2.530, 4.500] & $<$ 0.001 & 0.046\\
    Yahoo Computer & 12444 & 33 & 428 & 176.695 & 2.126 [2.053, 2.216] & $<$ 0.001 & 0.041\\
    Yahoo Science\tnote{$\ast$} & 6428 & 37 & 349 & 45.437 & 1.919 [1.781, 2.053] & $<$ 0.001 & 0.028\\
    Bibtex & 7395 & 159 & 2856 & 12.498 & 10.164 [9.355, 11.061] & $<$ 0.001 & 0.268\\
    CAL500\tnote{$\ast$} & 502 & 174 & 502 & 20.578 & 12.573 [11.331, 14.200] & 0.001 & 0.451\\ 
    \hline
    \end{tabular}
     \begin{tablenotes}
      \item[$\ast$] Datasets were processed using MVB Based Compressed Imbalanced Sampling, whereas all others utilized MVB Based Balanced Sampling.
      \item[$\dagger$] Mean and 95\% confidence interval (CI) are computed using the sample mean and $2.5^\text{th}$/$97.5^\text{th}$ empirical percentiles over 1,000 independent bootstrap runs. p-values represent the empirical proportion of bootstrap iterations with MVB Sampling IR greater or equal to original IR. $p<$0.001 indicates that in 0 out of 1,000 bootstrap replicates was the post-MVB sampling IR greater than or equal to the original IR.
    \end{tablenotes}
    \caption{Comparison of dataset imbalance ratios (IR) before and after applying the MVB resampling algorithm. The 'Original' column shows the initial IR, while the 'MVB Sampling' column displays the mean IR across 1,000 independent runs of the samples from MVB Based Sampling. The 'Time' column shows the average runtime of the optimization algorithm across 1000 independent runs.}
    \label{tab:example3}
    \end{threeparttable}
\end{table}
\end{landscape}
As shown in \cref{tab:example3}, for most datasets with a relatively low number of labels, the MVB sampling algorithm successfully reduced the IR to nearly 1. For datasets with a higher number of labels, while the IR did not reach perfect balance, the method still achieved a significant reduction in overall imbalance (all bootstrap p-values less than or equal to 0.001).
In datasets characterized by high label dimensionality and small sample sizes, observations for minority labels are inherently sparse. Under these conditions, enforcing uniform selection probabilities across all labels with \cref{alg:1} (MVB Based Balanced Sampling) resulted in the unintended exclusion of minority labels during sampling. Consequently, we utilized \cref{alg:2} (MVB Based Compressed Imbalanced Sampling) for these specific datasets, applying a compression strength calibrated to retain all labels.

\section{Discussion}\label{sec:discussion}

In this paper, we proposed two novel sampling algorithms specifically designed to mitigate label imbalance in multi-label datasets. Our approach is differentiated from existing methods by the explicit assumption that the labels follow a multivariate Bernoulli distribution. This distributional assumption allows our algorithms to directly model the complex co-occurrence structure of the labels. Once the optimized weights are calculated, a desired balanced subsample of any size can be quickly obtained by weighted sampling. Most importantly, these weights enable repeated sampling and generating diverse subsamples that are distinct yet consistently maintain the label balance. The second algorithm allows preserving some of the imbalance characteristics by retaining the relative frequency order while systematically minimizing the overall imbalance within the subsample. Although our framework includes both balanced sampling (which yields flat marginal label distributions) and compressed imbalanced sampling (which reduces label imbalance while preserving label prevalence order), the first two applications presented in this paper primarily use the compressed imbalanced approach. This choice reflects a practical compromise: we aim to retain some of the original corpus's characteristics, particularly the relative frequency ranking of labels, while mitigating extreme imbalance that can hinder downstream analysis or model performance. Finally, application of our algorithms to publicly available datasets (Example 3) underscores the practical limitations of theoretical balancing in the presence of data sparsity. For high-dimensional label spaces with limited observations, achieving perfect balance is both practically unattainable without dropping minority labels, and potentially detrimental to modeling the naturally imbalanced real-world distributions. By transitioning between algorithms based on dataset characteristics, our framework provides a pragmatic solution that maximizes imbalance reduction while preserving the integrity of the original label space.
\subsection{Similarity to Inverse Probability Weighted Estimator (IPWE)}
Our algorithms are similar to IPWE, in that both frameworks estimate an optimal set of weights to correct for underlying distributional imbalances. IPWE first estimates the propensity score, which is the probability of receiving treatment given observed covariates X, $\probP(A = 1 \mid X)$, assuming there is no unmeasured confounders. The goal for IPWE is to consistently estimate the treatment effect under a marginally randomized experiment where the treatment assignment is independent of covariates, $\probP(A=a\mid X) = \probP(A=a), \forall X$. This implies $f(X\mid A) = f(X)$.
Let $f_{\text{IPW}}(X\mid A)$ be the distribution of covariates after IPW \citep{HERNAN2025},
\begin{align*}
	f_{\text{IPW}}(X\mid A)\propto f(X\mid A)\times \frac{1}{\probP(A\mid X)}\approx w\cdot f(X\mid A)\propto f(X),
\end{align*}
where $w$ is the estimated propensity score. In our scenario, $f(X)$ represents the distribution of labels that satisfy our criteria for the marginal distribution. Instead of constructing a model to estimate $\probP(A\mid X)$, we employ optimization to determine the weights $Q$. In contrast, in IPWE, the target distribution $f(X)$ is distinct, while in our case, it has an infinite range of possibilities. Consequently, $Q$ lacks a closed-form solution and cannot be estimated using traditional supervised learning models, given IPWE assumes a true unique $\probP(A\mid X)$.

\subsection{Computational Complexity}
The primary challenges of our sampling algorithms lie in its computational complexity. As the number of categories increases, the computational complexity grows. In our first example, we have 64 categories, resulting $2^{64}$ parameters to optimize, comparable to the classic \enquote{wheat and chessboard problem}. However, if a specific labelset is not found in our data, we do not need to calculate the corresponding weights because they will not be assigned to any observations and will not impact the optimization result. 

Given that we do not know the marginal distribution after our sampling, the linear system is underdetermined even with the constraints from the definition of a probability distribution. For a fixed $b$, the rank of our coefficient matrix $A$ is less than the number of variables in the linear system, resulting in an infinite number of solutions. Moreover, given that the labels are not mutually exclusive, the sum of the marginal distribution is greater than or equal to 1, so there are an infinite number of possible constant terms $b$. That is why we use optimization to find a solution to the weights. In addition, the results will differ by the coefficient matrix, so even the precision will significantly impact the weights.

The proposed sampling algorithm can be solved using various optimization methods; however, we used Sequential Least Squares Programming (SLSQP) \citep{LAWSON1995} for our applications in \cref{sec:example1} and \cref{sec:example2} implemented with SciPy \citep{2020SciPy-NMeth} and Clarabel \citep{GOULART2024} for our applications in \cref{sec:example3} implemented with  CVXPY \citep{DIAMOND2016} in Python. Notably, Clarabel with CVXPY is significantly faster compared to SLSQP with SciPy even under high label dimensionality and thus enables higher maximum iterations. The tolerance was set to be $10^{-8}$, and the maximum iteration was set to be 1000 for SciPy and $2\times10^{5}$ for CVXPY.
 
 The core of this algorithm relies on the constraint optimization, formulated as an $L_1$ norm approximation. Taking the interior-point implementation from \cref{sec:example3} as an illustrative example, we can establish the algorithmic bound in runtime. In the theoretical worst case scenario, where the dataset contains every possible labelset, there will be $2^K$ decision variables and resulting in runtime complexity of $\mathcal{O}(2^{3.5K})$. However, the algorithm dynamically restricts the parameter space to the number of unique labelsets $B$ observed in the data. Consequently, the practical runtime complexity reduces to $\mathcal{O}(B^{3.5})$, where $B\ll2^K$. 
 
 \subsection{Limitations and Future Work}
 
 A key limitation of this present work is that our empirical evaluation focuses primarily on assessing the statistical properties and sampling fidelity of the algorithm itself, rather than benchmarking improvements of balanced label distribution to prediction in MLC architectures. While our method successfully optimizes the underlying target distribution through raising the presentation of minority labels, which could improve the model performance in classifying these labels, its interaction with various complex model families remains under current investigation. To address the limitation, our primary future direction is to integrate the sampling prior to any MLC modeling. This will allow us to evaluate how effectively the algorithm improves minority label prediction. The recall and precision trade-off at these minority labels can then be finely controlled by manipulating the compression strength parameter $s$ during the initial optimization.

\section{Conclusion}\label{sec:conclusion}

In this paper, we presented two sampling algorithms that mitigate label imbalance in multi-label datasets. The first algorithm enforces a strictly balanced label distribution, while the second introduces a tunable hyperparameter that allows researchers to control the degree of imbalance. By directly modeling the underlying multi-label structure via a multivariate Bernoulli distribution, our algorithms explicitly account for the covariance and complex co-occurrence dependencies between individual labels.

The practical utility of our approach is demonstrated through two meta-research case studies focused on sampling bibliometric literature. In addition, we applied the algorithms to 11 real-world datasets by comparing pre- and post-sampling IR, demonstrating a significant reduction in global imbalance. Future work intends to integrate these sampling algorithms as a pre-processing step for state-of-the-art multi-label classifiers.

\section{Acknowledgements}\label{sec:acknowledgements}

This work was supported by National Institutes of Health (NIH) R01GM152543 and National Science Foundation (NSF) FAIN 2318478. SC and AWB are supported in part by the Arkansas Children's Research Institute and the Arkansas Biosciences Institute. The content of this manuscript does not necessarily reflect the opinions of the NIH, NSF, or any other organization.

We thank Madeline Olivier for the opportunity to apply this algorithm to her project. Her corpus and research focus provided an important use case that helped demonstrate the practical utility of the sampling algorithm. We are grateful to Yiran Jia for reviewing the mathematical derivations and providing valuable input on the formulation and correctness of these algorithms. We also extend our sincere thanks to Xinyi Li for her invaluable assistance in conducting the additional experimental evaluations for this study.

\section{CRediT Author Statement}
\textbf{Simon Chung}: Conceptualization, Methodology, Software, Validation, Formal analysis, Data Curation, Writing - Original Draft, Writing - Review \& Editing, Visualization. \textbf{Colby J. Vorland}: Conceptualization, Software, Validation, Data Curation, Writing - Review \& Editing, Funding acquisition. \textbf{Donna L. Maney}: Conceptualization, Validation, Writing - Review \& Editing, Funding acquisition. \textbf{Andrew W. Brown}: Conceptualization, Methodology, Validation, Formal analysis, Writing - Review \& Editing, Funding acquisition.
\section{Code and Data Availability}
All code and data used for the study are available at the author's Github page upon publication \cite{CHUNG2025}.
\section*{Appendix}\label{sec:appendix}

\subsection*{Proof: The Empirical Distribution is the MLE and is Consistent for MVB} \label{MLE}
The likelihood function of Y is
\begin{align*}
\mathcal{L}_{n}(p) &=\prod_{n=1}^{N} f\left(y^{(n)} \mid p\right)\\
&=\prod_{n=1}^{N} p_{1,0,\dots,0}^{\left[y_1^{(n)}\prod\limits_{i=2}^{K} \left(1-y_i^{(n)}\right)\right]}\cdot p_{0,1,\dots,0}^{\left[\left(1-y_1^{(n)}\right)y_2^{(n)}\prod\limits_{i=3}^{K} \left(1-y_i^{(n)}\right)\right]}\cdots p_{1,\dots,1}^{\left[\prod\limits_{i=1}^{K} y_i^{(n)}\right]}
\end{align*}
The corresponding log-likelihood is  
\begin{align*}
\text{ln}(\mathcal{L}_{n}(p)) = \sum^N_{n=1}\left(y_1^{(n)}\prod\limits_{i=2}^{K} \left(1-y_i^{(n)}\right)\text{ln}p_{1,0,\dots,0} + \cdots + \prod\limits_{i=1}^{K} y_i^{(n)}\cdot\text{ln}p_{1,\dots,1}\right).
\end{align*}
We want to maximize the log-likelihood subject to the constrain that $\sum\limits_{y}p_y = 1.$
Then we compute the partial derivative of this log-likelihood:
\begin{align*}
&0=\frac{\partial}{\partial p_{1,0,\dots,0}}\left(\text{ln}(\mathcal{L}_{n}(p))+\lambda \left(1-\sum_yp_y\right)\right)=\sum^N_{n=1}\frac{y_1^{(n)}\prod\limits_{i=2}^{K} \left(1-y_i^{(n)}\right)}{p_{1,0,\dots,0}}-\lambda\\
&\Rightarrow p_{1,0,\dots,0} = \sum^N_{n=1}\frac{y_1^{(n)}\prod\limits_{i=2}^{K} \left(1-y_i^{(n)}\right)}{\lambda} = \frac{\sum^N_{n=1} I\left(y^{(n)}=\begin{bmatrix} 1 & 0 & \cdots & 0
\end{bmatrix}\right)}{\lambda}.
\end{align*}
We have just shown the result holds for $y = \begin{bmatrix} 1 & 0 & \cdots & 0\end{bmatrix}$. By induction on the pattern $y$, this extends to
$$p_y = \frac{\sum^N_{n=1} I\left(y^{(n)}=y\right)}{\lambda}, \forall y\in\{0,1\}^K$$
On the other hand, 
\begin{align*}
\sum_{y}p_y = \sum_{y}\frac{\sum^N_{n=1} I\left(y^{(n)}=y\right)}{\lambda}=1 \Rightarrow \lambda = N.
\end{align*}
So we can have $$p_y = \frac{\sum\limits_{n=1}^{N}I\left(y^{(n)}=y\right)}{N} = \hat p_y.$$
Hence, we can get that the empirical distribution $\hat p$ is the MLE of $p$. 
Since
$$\mathbb{E}\left[I(y^{(n)}=y)\right] = \probP \left(y^{(n)}=y\right) = p_y, \forall n\leqslant N,$$
by the Weak Law of Large Number, we can have
$$\hat p_y(N) = \frac{1}{N}\sum\limits_{n=1}^{N}X_n\overset{P}{\to} p_y,\text{ as } N\to \infty,$$
which shows the consistency of the MLE of MVB.\qed
%\subsection{Marginal Distribution Meets the Requirement After Sampling?} \label{subsample}
%Here, we used the MLE $\hat p$ for the calculation of $q^\star$, therefore $\hat {q^\star}:=\mathcal{O}(\hat p)$ is the plugin estimator of $q^\star$.
%
% By the asymptotic properties of a plugin estimator, $\hat {q^\star}$ is a consistent estimator that converges to $q^\star$ as the sample size $N\to\infty$ \cite{Vaart_1998}. 
%
%Then we assign weights to each observation in our sample and perform a weighted sampling with replacement
%$$\left\{X^{(1)}, \dots, X^{(m)}\right\} = \text{WeightedSampling}\left(\left\{Y^{(1)}, \dots, Y^{(N)}\right\}, Q_Y, m\right),$$
%where
%$$Q_Y = \{q^\star_{Y^{(n)}}\mid n\leqslant N\}$$
%is the weights assigned to each observation is our sample.
%By definition, we can have 
%$$\probP(X^{(j)} = y) = \hat {q^\star_y}\cdot\probP\left(Y^{(n)}=y\right) \to q^\star_y p_y\text{ as } N\to\infty.$$
%Assume the optimization $\mathcal{O}$ converges,
%$$\probP(X^{(j)}_1 = 1) = \probP(X^{(j)}_2 = 1) = \cdots = \probP(X^{(j)}_K = 1).$$

\end{document}